# LegalBench-BR: A Benchmark for Evaluating Large Language Models on Brazilian Legal Decision Classification

Pedro Barbosa de Carvalho Neto

Teresina, PI, Brazil

**Abstract**

We introduce LegalBench-BR, the first public benchmark for evaluating language models on Brazilian legal text classification. The dataset comprises 3,105 appellate proceedings from the Santa Catarina State Court (TJSC), collected via the DataJud API (CNJ) and annotated across five legal areas through LLM-assisted labeling with heuristic validation. On a class-balanced test set, BERTimbau-LoRA — updating only 0.3% of model parameters — achieves **87.6% accuracy and 0.87 F1 macro** (+22pp over Claude 3.5 Haiku, +28pp over GPT-4o mini). The gap is most striking on *administrativo* (administrative law): GPT-4o mini scores F1 = 0.00 and Claude Haiku scores F1 = 0.08 on this class, while the fine-tuned model reaches F1 = 0.91. Both commercial LLMs exhibit a systematic bias toward *cível* (civil law), absorbing ambiguous classes rather than discriminating them — a failure mode that domain-adapted fine-tuning eliminates. These results demonstrate that general-purpose LLMs cannot substitute for domain-adapted models in Brazilian legal classification, even when the task is a simple 5-class problem, and that LoRA fine-tuning on a consumer GPU closes the gap at zero marginal inference cost. We release the full dataset, model, and pipeline to enable reproducible research in Portuguese legal NLP.

## 1. Introduction

The intersection of Natural Language Processing (NLP) and legal text analysis has seen significant advances in English-language contexts, with benchmarks such as LegalBench (Guha et al., 2023) providing standardized evaluation frameworks for 162 legal reasoning tasks. However, Portuguese-language legal NLP — particularly for the Brazilian judicial system — remains underserved despite Brazil operating one of the largest court systems in the world, with over 80 million active cases across 91 courts.

This gap is not merely academic. Brazilian law firms, legal tech startups, and the judiciary itself increasingly need automated classification of proceedings for case routing, workload distribution, and analytics. The Brazilian judiciary's DataJud system, maintained by the National Council of Justice (CNJ), provides public API access to structured data from all state courts — yet no standardized benchmark exists to evaluate how well language models perform on this data.

A natural question arises: do general-purpose LLMs, with their broad multilingual training, already perform well enough on Brazilian legal classification to obviate the need for domain-specific models? Our benchmark provides an empirical answer.

**Contributions:**

— We release **LegalBench-BR**, the first public Portuguese legal NLP benchmark with balanced class representation across 5 legal areas, available on HuggingFace.

— We demonstrate a **+22pp accuracy gap** between a lightweight fine-tuned model (BERTimbau + LoRA) and the best commercial LLM (Claude 3.5 Haiku), with the gap reaching +91pp F1 on the hardest class.

— We identify a **systematic "cível default" bias** in zero-shot LLM classification of Brazilian legal text, where both Claude and GPT absorb ambiguous classes into the majority category.

— We document a **complete, reproducible pipeline** — from DataJud API collection through targeted rebalancing, annotation, training, and evaluation — that can be adapted for other Brazilian courts.

## 2. Related Work

**English Legal NLP Benchmarks.** LegalBench (Guha et al., 2023) established 162 tasks for evaluating LLMs on legal reasoning across six categories. CUAD (Hendrycks et al., 2021) targets contract understanding with expert annotations. CaseHOLD (Zheng et al., 2021) tests legal holding identification. These benchmarks have driven significant progress but are limited to English and common-law jurisdictions, with no equivalent for civil-law systems like Brazil's.

**Portuguese NLP Models.** BERTimbau (Souza et al., 2020) provides BERT models pre-trained on the BrWaC corpus of Brazilian Portuguese. Albertina-PT (Rodrigues et al., 2023) extends the DeBERTa architecture to Portuguese. These models provide strong foundations for downstream tasks but have not been systematically evaluated on legal classification.

**Brazilian Legal NLP.** LeNER-BR (Luz de Araujo et al., 2018) offers named entity recognition for Brazilian legal texts. UlyssesNER-BR (Albuquerque et al., 2022) extends this to legislative documents. Victor (Lage-Freitas et al., 2022) classifies appeal themes at the Brazilian Supreme Court (STF). However, no work addresses multi-class legal area classification at the state appellate level with a public, reproducible benchmark and LLM baselines.

**LoRA and Parameter-Efficient Fine-Tuning.** LoRA (Hu et al., 2021) enables fine-tuning large language models by training low-rank decomposition matrices, typically updating < 1% of parameters. This approach has been shown effective for domain adaptation in specialized fields including biomedical NLP (Chen et al., 2023) and financial text (Yang et al., 2023), but its application to Portuguese legal text classification has not been benchmarked.

## 3. Dataset Construction

### 3.1 Data Collection

We collected appellate proceedings from TJSC (Tribunal de Justiça de Santa Catarina) using the DataJud public API, an ElasticSearch-based endpoint providing structured metadata on Brazilian court cases.

**Initial collection** yielded 2,000 proceedings via diversified queries across procedural classes. Analysis revealed severe class imbalance: 97% of results originated from *Câmaras de Direito Comercial* (Commercial Law chambers), producing a dataset dominated by civil/commercial law.

**Targeted collection** addressed this imbalance by querying specifically by court chamber type (*Câmaras Criminais*, *Câmaras de Direito Civil*, *Varas da Fazenda Pública*) and subject matter keywords (e.g., "execução fiscal," "habeas corpus," "mandado de segurança"), adding 1,454 proceedings with representation across underserved legal areas.

**Data fields per record:** procedural class (*classe*), subject matter(s) (*assuntos*), court chamber (*órgão julgador*), up to 20 procedural movements with dates and annotations, and filing date. CPF and CNPJ identifiers were anonymized via regex substitution prior to storage.

### 3.2 Annotation Pipeline

Annotation followed a two-stage process designed to combine LLM scalability with rule-based precision:

**Stage 1 — LLM classification.** Claude 3.5 Haiku classified each record via structured JSON prompts incorporating procedural class, subject matter, and court chamber context. The model returned a legal area label, confidence score (0–1), and one-sentence justification. Records with confidence < 0.7 were discarded (573 of 3,678, or 15.6%).

**Stage 2 — Heuristic correction.** Rule-based overrides corrected 196 misclassifications (6.3% of retained records) using high-confidence metadata signals:

— *Court chamber rules:* "Câmara Criminal" → penal, "Câmara de Família" → família (confidence: 0.95)

— *Procedural class rules:* "Apelação Criminal" → penal, "Execução Fiscal" → tributário (confidence: 0.92)

— *Subject matter rules:* "Tráfico de Drogas" → penal, "ICMS/IPTU" → tributário (confidence: 0.88)

This two-stage approach addresses a well-documented weakness of LLM annotation: the tendency to default to majority classes when discriminating signals are subtle (Pangakis et al., 2023).

### 3.3 Label Set Design

Analysis of the annotated corpus informed principled label set reduction:

— **Trabalhista** (labor law) was excluded: TJSC is a state court; labor cases are constitutionally assigned to the *Tribunal Regional do Trabalho* (TRT-12), a separate jurisdiction. Zero labor cases appeared in the corpus.

— **Família** (family law) was merged into *cível*: only 45 appellate-level examples existed, as Brazilian family law cases are predominantly resolved at first instance (*Varas de Família*).

**Final label set (5 classes):** cível (civil), consumidor (consumer), tributário (tax), administrativo (administrative), penal (criminal).

### 3.4 Class Balancing

The natural distribution of the corpus is heavily skewed (cível: 52%, penal: 18%, administrativo: 15%, consumidor: 8%, tributário: 7%). To produce a benchmark where majority-class prediction is not a viable strategy, we applied:

— **Train:** oversampling of minority classes to 363 examples per class (1,815 total), ensuring equal gradient contribution per class during training.

— **Validation / Test:** undersampling of majority classes to 21 examples per class (105 each), ensuring F1 macro is a meaningful and interpretable metric.

Random baseline accuracy: 20%. Random baseline F1 macro: 0.20.

### 3.5 Dataset Summary

| Split | Total | Per Class | Method | Purpose |
| --- | --- | --- | --- | --- |
| Train | 1,815 | 363 | Oversampling | Model training |
| Validation | 105 | 21 | Undersampling | Hyperparam. selection |

| Test | 105 | 21 | Undersampling | Final evaluation |

Source corpus: 3,678 proceedings collected, 3,105 retained after confidence filtering.

## 4. Experiments

### 4.1 Models

**BERTimbau + LoRA.** Base model: `neuralmind/bert-base-portuguese-cased` (110M parameters). LoRA configuration: rank r=16, scaling α=32, target modules: query and value attention layers, dropout 0.1. Total trainable parameters: 0.3% of the base model. Training: 5 epochs on Google Colab T4 GPU (16GB VRAM), learning rate 2e-4 with linear warmup, batch size 16, FP16 mixed precision, best checkpoint selected by validation F1 macro.

**Claude 3.5 Haiku.** Accessed via OpenRouter API. Zero-shot classification with a standardized prompt: *"Classifique esta descrição de processo judicial brasileiro em uma das seguintes áreas do direito: cível, consumidor, tributário, administrativo, penal. Responda com UMA ÚNICA PALAVRA."* Temperature: default. Max tokens: 20.

**GPT-4o mini.** Accessed via OpenAI API. Same standardized prompt and constraints as Claude Haiku, with an additional system message: *"Você é um classificador de processos judiciais brasileiros. Responda com uma única palavra."*

### 4.2 Evaluation Protocol

All three models were evaluated on the identical class-balanced test set (105 samples, 21 per class). Metrics: accuracy, macro-averaged precision, recall, and F1, per-class precision/recall/F1, and average inference latency (ms per sample, measured end-to-end including network round-trip for API models).

## 5. Results

### 5.1 Overall Performance

| Model | Accuracy | F1 Macro | Latency (ms) | Cost / 1k |
|---|---|---|---|---|
| **BERTimbau-LoRA** | **87.6%** | **0.87** | **233** | ~$0.00 |
| Claude 3.5 Haiku | 65.7% | 0.63 | 1,444 | ~$0.15 |
| GPT-4o mini | 60.0% | 0.57 | 964 | ~$0.10 |

BERTimbau-LoRA outperforms both commercial LLMs by a wide margin: +22 percentage points over Claude Haiku and +28pp over GPT-4o mini in accuracy. The F1 macro gap (+0.24 and +0.30, respectively) confirms this advantage is not driven by a single class.

### 5.2 Per-Class Performance

| Class | BERTimbau F1 | Claude F1 | GPT F1 | Δ Best LLM |
|---|---|---|---|---|
| Administrativo | **0.91** | 0.08 | 0.00 | **+0.83** |
| Cível | **0.72** | 0.53 | 0.47 | **+0.19** |
| Consumidor | 0.81 | **0.84** | 0.58 | −0.03 |
| Penal | **0.98** | 0.92 | 0.92 | **+0.06** |
| Tributário | **0.95** | 0.79 | 0.86 | **+0.09** |

The most striking result is on *administrativo*: GPT-4o mini achieves **F1 = 0.00** (zero correct predictions) and Claude Haiku achieves F1 = 0.08, while BERTimbau-LoRA reaches F1 = 0.91. This 83-point gap on a single class reveals a fundamental limitation of zero-shot LLM classification on domain-specific legal categories.

The only class where an LLM matches the fine-tuned model is *consumidor*, where Claude Haiku slightly outperforms BERTimbau (0.84 vs. 0.81). This is consistent with consumer law's relatively distinctive vocabulary ("vício do produto," "inclusão indevida em cadastro"), which zero-shot models can leverage.

### 5.3 Error Analysis

**The "cível default" bias.** Both LLMs exhibit a systematic pattern: when uncertain, they classify cases as *cível*. Claude Haiku assigns cível to 38% of test samples (true proportion: 20%), achieving 81% recall but only 40% precision on that class. GPT-4o mini shows the same pattern (81% recall, 33% precision). This mirrors the statistical distribution of Brazilian court caseloads, where civil cases are the plurality — suggesting that LLMs have internalized corpus-level frequency distributions rather than learning discriminative features.

**Administrativo vs. cível confusion.** Administrative law cases in Brazil (mandado de segurança, ação civil pública, ação popular) share procedural class names with civil cases and often involve the same courts. Zero-shot models lack the contextual signal that a case heard by a *Vara da Fazenda Pública* or involving *servidor público* subject matter is administrative rather than civil. The fine-tuned model learns these signals from training data.

**Penal as a discriminability ceiling.** Criminal cases use distinctive procedural classes ("Apelação Criminal," "Habeas Corpus Criminal"), distinctive vocabulary ("pena privativa de liberdade," "tráfico de drogas"), and are heard in dedicated *Câmaras Criminais*. All three models achieve F1 ≥ 0.92, confirming that high discriminability yields strong zero-shot performance.

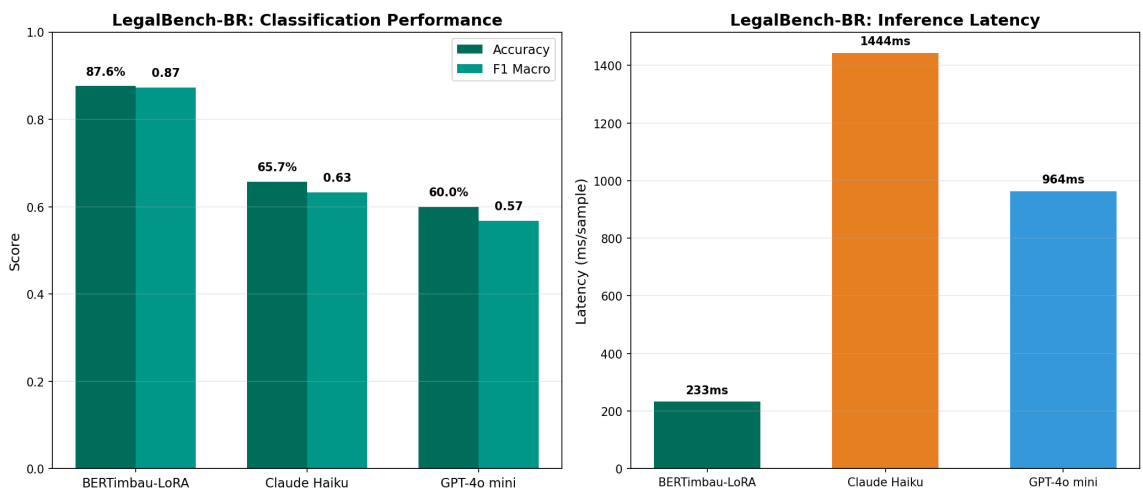

*Figure 1: Classification performance (left) and inference latency (right) across models. BERTimbau-LoRA dominates on both metrics.*

### 5.4 Discussion

*Practical Implications for Legal Technology*

**Deployment economics at scale.** A mid-size law firm processing 10,000 cases per month would spend approximately $150/month using Claude Haiku for classification, with each request taking ~1.4 seconds round-trip. BERTimbau-LoRA runs at 233ms per sample on a CPU, costs $0 in inference, and — critically — **does not transmit case data to external servers**. This last point is a regulatory requirement: Brazilian bar association (OAB) ethics rules and the LGPD (Lei Geral de Proteção de Dados) impose strict obligations on confidentiality and data minimization that are difficult to satisfy when routing case text through third-party APIs.

**The "cível default" bias is not benign.** In a production system, systematically misclassifying administrative and tax cases as civil would cause downstream failures: incorrect case routing, wrong workflow assignments, and flawed caseload analytics. The fact that both LLMs independently converge on this bias — despite different training corpora — suggests it is a structural property of how these models represent legal Portuguese, not a model-specific artifact. Domain fine-tuning breaks this bias explicitly through balanced training data and exposure to discriminative metadata features.

**Transferability to other Brazilian courts.** TJSC was selected for API accessibility, but the five legal areas in our benchmark correspond to divisions present in all 27 Brazilian state courts. The procedural classes (apelação cível, apelação criminal, mandado de segurança, execução fiscal) are defined by federal legislation (CPC and CPP) and are uniform nationwide. We expect the fine-tuned model to generalize to other state courts with minimal performance degradation, though this requires empirical validation.

*Methodological Implications*

**Balanced evaluation changes the narrative.** On the original unbalanced test set (78% cível), Claude Haiku achieved 94.5% accuracy — a number that would suggest LLMs are already "good enough." After balancing to 20% per class, accuracy dropped to 65.7%. This 29-point difference is entirely attributable to the removal of the cível-majority shortcut, and underscores the importance of balanced benchmarks in legal NLP, where class distributions are naturally skewed and majority-class performance is uninformative.

**LLM annotation requires validation.** Our annotation pipeline revealed that Claude Haiku itself systematically misclassified 6.3% of records — predominantly defaulting administrative cases to cível. This is the same bias we later observed in the benchmark evaluation. Researchers using LLM-assisted annotation for legal datasets should implement heuristic validation layers, particularly when high-confidence metadata (court chamber, procedural class) is available.

## 6. Conclusion

We presented LegalBench-BR, the first public benchmark for Brazilian legal text classification, addressing a critical gap in non-English legal NLP. Our central finding is that a lightweight fine-tuned model (BERTimbau + LoRA, 0.3% trainable parameters, trained in 30 minutes on a free Colab GPU) substantially outperforms commercial LLMs on domain-specific classification: 87.6% accuracy vs. 65.7% for Claude 3.5 Haiku and 60.0% for GPT-4o mini.

The most consequential result is the failure mode we identified: both LLMs exhibit a systematic "cível default" bias that renders them effectively unusable for administrative law classification (F1 ≤ 0.08). This finding has immediate practical implications for legal technology companies building classification pipelines on top of general-purpose LLMs, and suggests that domain-specific fine-tuning remains essential for specialized legal NLP tasks, even in the era of large language models.

**Limitations:**

— Dataset is limited to TJSC appellate decisions; generalization to other courts requires validation.

— Automated annotation may carry forward Claude Haiku's classification biases despite heuristic correction.

— Label set excludes labor law (separate jurisdiction) and family law (insufficient appellate data).

— Test set size (105 samples) provides limited statistical power for per-class confidence intervals.

— BERTimbau-LoRA's advantage may partly reflect information leakage from metadata features (court chamber names) that LLMs see in the text but may not leverage as effectively.

**Future work:**

— Expand to multi-court evaluation (TJSP, TJRJ, STJ) to test cross-court generalization.

— Add tasks of increasing complexity: case outcome prediction, legal summarization, entity extraction.

— Benchmark larger Portuguese models (Albertina-PT-BR, Sabiá) with and without LoRA.

— Evaluate few-shot and chain-of-thought prompting strategies to improve LLM baselines.

— Include full-text decisions (currently limited to metadata) when available via court APIs.

**Reproducibility.**

All code, data, and trained model weights are publicly available. **Dataset:** HuggingFace `pedronettotrue/legal-nlp-benchmark-br`. **Model:** HuggingFace `pedronettotrue/bertimbau-legal-tjsc`.